\documentclass[10pt]{article} 
\usepackage[preprint]{tmlr}

\usepackage{booktabs}
\usepackage{graphicx}
\usepackage{caption}
\usepackage{subcaption}
\usepackage{amsmath}

\usepackage[linesnumbered, ruled, vlined]{algorithm2e}

\SetCommentSty{mycommfont}
\SetKwInput{KwInput}{Input}                
\SetKwInput{KwOutput}{Output} 
\SetKwInput{KwParams}{Parameters}  

\usepackage{amsthm}
\usepackage{thmtools} 
\theoremstyle{definition}
\newtheorem{definition}{Definition}[section]
\newtheorem{theorem}{Theorem}[section]

\usepackage{amssymb}
\usepackage{mathabx}
\usepackage{hyperref}
\hypersetup{
     colorlinks   = true,
     linkcolor    = blue,
     citecolor    = blue
}

\newcommand{\rBr}[1]{\left(#1\right)}
\newcommand{\sqBr}[1]{\left[#1\right]}

\newcommand{\vecOp}[1]{\operatorname{vec}\left(#1\right)}

\newcommand{\Real}{\mathbb{R}}
\newcommand{\tens}[1]{\boldsymbol{\mathcal{#1}}} 
\newcommand{\mat}[1]{\boldsymbol{#1}} 
\newcommand{\vect}[1]{\boldsymbol{#1}} 
\newcommand{\inRe}[1]{\in\Real^{#1}} 

\usepackage{hyperref}
\usepackage{url}

\title{Gradient Boosting for Efficient Rank Discovery \\ in Tensor Network Kernel Machines}

\author{\name Albert Saiapin \email a.saiapin@tudelft.nl \\
      \addr Delft Center for Systems and Control\\
      Delft University of Technology
      \AND
      \name Kim Batselier \email k.batselier@tudelft.nl \\
      \addr Delft Center for Systems and Control\\
      Delft University of Technology
      }

\def\month{MM}  
\def\year{YYYY} 
\def\openreview{\url{https://openreview.net/forum?id=XXXX}} 

\begin{document}
\maketitle

\begin{abstract}
Tensor network kernel machines (TNKMs) combine nonlinear features with low-rank tensor representations to model complex relationships efficiently. However, selecting an appropriate tensor rank remains challenging, as conventional rank-search methods require training multiple candidate models, increasing computational costs. In this work, we focus on a canonical polyadic (CP) tensor network parameterization and propose GB-CPKM, a gradient boosting framework that builds a higher-capacity predictor by sequentially adding fixed-rank learners. 
This approach provides a new perspective on rank selection by incrementally increasing model capacity, with ensemble size selected based on validation performance rather than through explicit rank search. 
Experiments on real-world regression benchmarks show that GB-CPKM trains up to $6\times$ faster and uses up to $2\times$ less peak memory than a conventional rank-selection method based on cross-validation, while achieving comparable predictive performance. These results highlight the potential of gradient boosting to improve the computational efficiency and scalability of tensor-network learning.
\end{abstract}
\section{Introduction}
Machine learning (ML) has become an essential tool in a wide range of scientific and industrial applications, from image classification~\citep{drl_he_2016} and natural language processing~\citep{attention_vaswani_2017} to industrial control systems~\citep{drl_duan_2016}. However, achieving strong predictive performance requires more than fitting a model to training data. In addition to learning model parameters, practitioners must select appropriate hyperparameters~\citep{hpo_tong_2020}, which govern aspects of model architecture, complexity, training, and can substantially affect generalization performance and computational cost. Yet, hyperparameter selection often relies on grid search or other trial-and-error procedures~\citep{hpoML_franceschi_2025} that require training and evaluating multiple candidate models.

Among the hyperparameters that influence machine learning performance, model complexity is particularly important in methods based on low-rank tensor representations. Tensor networks (TNs) represent high-dimensional parameters using interconnected lower-dimensional tensors, potentially reducing storage and computational costs~\citep{TD_Kolda_2009,TD_Cichocki_2014}. They have consequently been applied to neural network compression and other machine learning tasks~\citep{TN1_Cichocki_2016,tnApp_cichocki_2017}. However, the benefits of these representations depend on the choice of tensor rank, which governs the trade-off between model expressiveness and computational cost. Selecting an appropriate rank is therefore a key challenge in tensor-network learning.

Tensor network kernel machines (TNKMs)~\citep{fftd_wesel_2021,lrtdnsi_2022_batselier} combine nonlinear feature mappings with low-rank tensor representations to learn in high-dimensional feature spaces. In canonical polyadic tensor network kernel machines (CPKMs)~\citep{tnkm_2026_saiapin}, the CP rank controls model capacity and influences computational and memory requirements.
Conventional rank selection often involves training and evaluating models at multiple candidate ranks, which can be computationally expensive. To address this challenge, we propose GB-CPKM, a gradient boosting framework that incrementally builds an ensemble of fixed-rank CPKM base learners. By adding learners sequentially, GB-CPKM increases model capacity and allows the ensemble size to be selected using validation performance, providing an alternative to explicit rank search.
The main contributions of this work are summarized as follows:
\begin{itemize}
    \item We introduce GB-CPKM, a gradient boosting framework that combines CPKM base learners to construct an expressive ensemble, enabling incremental model growth and effective rank discovery.
    \item We systematically analyze the computational and memory requirements of the proposed approach and investigate how base-learner rank and the number of boosting stages affect 
    trade-offs between model complexity, predictive performance, and computational efficiency.
    \item We evaluate GB-CPKM on regression benchmarks of varying sizes, comparing its performance, training time, and peak memory consumption against a cross-validation-based rank-selection baseline. The results show that GB-CPKM trains up to $6\times$ faster and uses up to $2\times$ less peak memory on average, while achieving comparable predictive performance. 
    \item We introduce GB-CPKM(H), a hybrid training strategy that combines the efficient rank discovery of gradient boosting with an additional training step over the full ensemble model. GB-CPKM(H) improves predictive accuracy while preserving the efficiency of boosting, making it a promising default choice for CP-based kernel models when both accuracy and computational cost matter.
\end{itemize}
\section{Tensor Network Kernel Machines}

\subsection{Notation}
We first establish the notation used throughout the paper. Calligraphic uppercase letters, such as $\mathcal{Z}$, denote sets. Scalars are denoted by italic letters (e.g., $w$ and $W$), vectors by lowercase bold letters (e.g., $\vect{w}$), matrices by uppercase bold letters (e.g., $\mat{W}$), and tensors by uppercase bold italic letters (e.g., $\tens{W}$). Indexed vectors are denoted by $\vect{w}_r^{(d)}$, where the indices and superscripts distinguish the corresponding vectors.

For a vector $\vect{w} \in \mathbb{R}^{I}$, its $i$th entry is denoted by $w_i$. Similarly, the $(i_1,i_2,\ldots,i_D)$th entry of a $D$th-order tensor
$\tens{W} \in \mathbb{R}^{I_1 \times I_2 \times \cdots \times I_D}$
is denoted by $w_{i_1 i_2 \cdots i_D}$. Unless stated otherwise, we use zero-based indexing for tensor entries. We use $\otimes$ to denote the Kronecker product~\citep{TN1_Cichocki_2016}. The transpose of a vector $\vect{w}$ is denoted by $\vect{w}^{\top}$.
For convenience, we represent tensors as vectors using the following vectorization convention.

\begin{definition}[Vectorization]
\label{def_vec}
The vectorization operator
\[
\vecOp{\cdot}:
\mathbb{R}^{I_1 \times I_2 \times \cdots \times I_D}
\rightarrow
\mathbb{R}^{I_1 I_2 \cdots I_D}
\]
maps a $D$th-order tensor $\tens{W}$ to a vector whose entries are the tensor entries arranged in a fixed order. Specifically,
\begin{equation*}
\vecOp{\tens{W}}_i
\coloneq w_{i_1 i_2 \cdots i_D},
\end{equation*}
where
\begin{equation*}
i \coloneq \rBr{i_1 i_2 \cdots i_D}
= i_1 + \sum_{d=2}^{D} i_d
\prod_{j=1}^{d-1} I_j.
\end{equation*}
Thus, the first tensor index varies fastest in the vectorized representation.
\end{definition}

\subsection{Tensor Networks}
Tensor networks (TNs) provide a framework for efficiently representing and manipulating high-dimensional tensors by expressing them as networks of smaller tensors connected through shared indices~\citep{TD_Cichocki_2014}. This representation can substantially reduce storage requirements when the underlying tensor admits a low-rank decomposition. In this work, we focus on the canonical polyadic decomposition.
\begin{definition}[Canonical polyadic decomposition (CPD)~\citep{TD_Kolda_2009}]
\label{CPD_Kron}
A $D$th-order tensor $\tens{W} \in \mathbb{R}^{I_1 \times I_2 \times \cdots \times I_D}$ admits a rank-$R$ CP decomposition if it can be expressed as
\begin{equation}\label{cpd_def}
    \vecOp{\tens{W}}
    \coloneq
    \sum_{r=1}^{R}
    \vect{v}^{(D)}_{r}
    \otimes \cdots \otimes
    \vect{v}^{(1)}_{r},
\end{equation}
where $\vect{v}^{(d)}_{r} \in \mathbb{R}^{I_d}$ and
\begin{equation*}
    \mat{V}^{(d)}
    \coloneq
    \sqBr{\vect{v}^{(d)}_{1}, \ldots, \vect{v}^{(d)}_{R}}
    \in \mathbb{R}^{I_d \times R}
\end{equation*}
is the factor matrix (core) associated with mode $d$.
\end{definition}

For a tensor with equal mode sizes, $I_d=I$, the original dense representation requires $\mathcal{O}(I^D)$ storage, whereas its CP representation requires only $\mathcal{O}(DIR)$ storage. Thus, when $R$ is sufficiently small relative to the mode sizes and tensor order, CPD can substantially reduce memory requirements. Another practical advantage is that CPD is characterized by a single rank parameter, $R$, which simplifies model selection. By contrast, tensor-train (TT)~\citep{TT_Oseledets_2011} and hierarchical Tucker (HT)~\citep{HT_Grasedyck_2010} decompositions involve multiple rank parameters, making their rank selection more involved.
CPD can also offer more favorable storage scaling than the Tucker decomposition~\citep{TD_Kolda_2009}. For equal mode sizes $I$ and a common Tucker rank $R$, the Tucker representation requires $\mathcal{O}(DIR + R^D)$ storage, owing to its factor matrices and core tensor. 

Despite its favorable storage properties, the effectiveness of CPD depends critically on the choice of rank $R$. A rank that is too low may limit model expressiveness, whereas an excessively high rank increases computational and memory costs and may lead to overfitting. Selecting an appropriate rank is therefore essential for balancing predictive accuracy and model complexity. However, conventional approaches typically evaluate multiple candidate ranks~\citep{TD_Kolda_2009}, requiring repeated model training and increasing computational cost. This motivates the development of more efficient rank-selection strategies.

\subsection{Model Formulation and Optimization}
Consider the following linear model:
\begin{equation*}
\label{intro_simple_model}
    f(\vect{x}) = \vect{\phi}(\vect{x})^\top \vect{w},
\end{equation*}
where $\vect{\phi}(\vect{x}) \in \mathbb{R}^{I_1 I_2 \cdots I_D}$ is a nonlinear feature map and $\vect{w} \in \mathbb{R}^{I_1 I_2 \cdots I_D}$ is a weight vector learned from the dataset $\mathcal{D} = \{\mat{X}\in \mathbb{R}^{N\times D}, \vect{y}\in \mathbb{R}^{N}\}$.

Tensor network kernel machines (TNKMs) extend this model by representing the weight vector as a tensor network and assuming that the feature map factorizes across the input dimensions:
\begin{equation*}
\label{tensor_product_features}
    \vect{\phi}(\vect{x})
    \coloneq
    \vect{\phi}^{(D)}(x_D) \otimes \cdots \otimes
    \vect{\phi}^{(1)}(x_1),
\end{equation*}
where $\vect{\phi}^{(d)}: \mathbb{R} \rightarrow \mathbb{R}^{I_d}$ maps the $d$th component of $\vect{x} \inRe{D}$ to a feature vector. This tensor-product construction is related to product kernels~\citep{VFF_Hensman_2017, GPRR_Solin_2019}.
Although the resulting feature space has dimension $\prod_{d=1}^{D} I_d$, which grows exponentially with the input dimension when the mode sizes are comparable, the tensor network representation of $\vect{w}$ enables model responses and gradients to be computed without explicitly constructing or storing the full weight vector. For the CP decomposition~\eqref{cpd_def}, this computational advantage is formalized by the following result.

\begin{theorem}[CP kernel machine~\citep{QTNM_Wesel_2024}]
\label{CPD_km}
Suppose that $\vect{w}$ admits a rank-$R$ CP decomposition. Then, for the model
\begin{equation*}
    f(\vect{x}) =
    \left[
    \vect{\phi}^{(D)}(x_D) \otimes \cdots \otimes
    \vect{\phi}^{(1)}(x_1)
    \right]^\top \vect{w},
\end{equation*}
its response and gradients can be computed in
$\mathcal{O}(DIR)$ time, rather than
$\mathcal{O}\left(\prod_{d=1}^{D} I_d\right)$, where
$I = \max_{d} I_d$.
\end{theorem}

Training a CPKM involves solving the following nonlinear, nonconvex optimization problem:
\begin{equation}
\label{cpd_km_opt}
\begin{split}
    \min_{\vect{v}}\quad &
    \frac{1}{2}
    \left\|\vect{y} - \mat{\Phi}\vect{w}(\vect{v})\right\|_2^2
    + \frac{\gamma}{2}
    \left\|\vect{v}\right\|_2^2, \\
    \text{subject to}\quad &
    \vect{w}(\vect{v}) =
    \sum_{r=1}^{R}
    \vect{v}^{(D)}_{r} \otimes \cdots \otimes
    \vect{v}^{(1)}_{r},
\end{split}
\end{equation}
where $\vect{y}$ contains the target values, $\mat{\Phi}$ is the feature matrix with entries $\Phi_{ni} \coloneq \phi_i(\vect{x}_n)$, and $\vect{x}_n$ is the $n$th input sample from the data matrix $\mat{X} \inRe{N\times D}$. The vector $\vect{v} \inRe{R\sum_d^D I_d}$ collects all entries of the cores $\mat{V}^{(d)}$, whose columns are denoted by $\vect{v}^{(d)}_{r}$. 
The parameter $\gamma$ controls the strength of the $\ell_2$ regularization. 

In this work, we focus on alternating least squares (ALS)~\citep{ALS_Uschmajew_2012}, which optimizes the tensor factors $\mat{V}^{(d)}$ iteratively, for $d=1, \dots, D$, while keeping the remaining factors fixed, providing a natural approach for exploiting the structure of CP decompositions.
\section{Gradient Boosting Tensor Network Kernel Machines}
Boosting is an ensemble learning technique that constructs a predictive model by combining multiple base learners into a stronger model. We construct the ensemble as an additive model~\citep{esl_hastie_2009} of the form
\begin{equation}
\label{gb_ensemble}
    f_M(\vect{x})
    =
    \sum_{m=1}^{M} \beta_m h_m(\vect{x}),
\end{equation}
where $h_m$ is the $m$th base learner and $\beta_m$ is its additive weight.
Rather than training all learners independently, boosting adds them sequentially, with each new learner aiming to correct the errors of the current ensemble~\citep{boosting_schapire_2012}. This sequential construction provides a flexible way to increase model capacity while controlling the complexity of individual learners. In the following, we introduce the main principles of gradient boosting and describe how they can be applied to tensor network kernel machines (see Theorem~\ref{CPD_km}).

\subsection{Gradient Boosting}
Gradient boosting (GB) provides a general framework for constructing additive models~\eqref{gb_ensemble}, using arbitrary differentiable loss functions~\citep{gfa_friedman_2001}. Given training data $\mathcal{D} = \{\rBr{\vect{x}_i, y_i} | \vect{x}_i \inRe{D}, y_i \inRe{},  i=1, \dots, N\}$, the objective is to minimize the empirical risk $\mathcal{L}(f) =
    \sum_{i=1}^{N} l\bigl(y_i, f(\vect{x}_i)\bigr)$,
where $l(y_i,f(\vect{x}_i))$ measures the prediction error for sample $i$.

Starting from an initial model $f_0$, gradient boosting builds an additive model by sequentially introducing new learners. At iteration $m$, it computes the gradient of the loss with respect to the current predictions:
\begin{equation*}
    g_{im}
    =
    \left.
    \frac{\partial l(y_i,\hat{y})}{\partial \hat{y}}
    \right|_{\hat{y}=f_{m-1}(\vect{x}_i)},
    \qquad i=1,\ldots,N.
\end{equation*}
The negative gradient, $-g_{im}$, indicates the direction in which the prediction for sample $i$ should move to decrease the loss locally. In function space, a gradient descent step would update the predictions on the training data as
\begin{equation*}
    f_m(\vect{x}_i)
    =
    f_{m-1}(\vect{x}_i) - \beta_m g_{im},
\end{equation*}
where $\beta_m$ is the step size.

However, this update specifies values only at the training samples and does not directly define a predictive function for unseen inputs. Gradient boosting addresses this by fitting a new learner $h_m$ to approximate the negative gradient from the input features:
\begin{equation*}
    h_m
    =
    \arg\min_{h \in \mathcal{H}}
    \sum_{i=1}^{N}
    \left[-g_{im} - h(\vect{x}_i)\right]^2,
\end{equation*}
where $\mathcal{H}$ denotes the class of candidate base learners. The resulting learner approximates the desired descent direction and can therefore be evaluated at both training and unseen inputs.
The ensemble is then updated according to
\begin{equation}\label{gb_update}
    f_m(\vect{x})
    =
    f_{m-1}(\vect{x}) + \beta_m h_m(\vect{x}).
\end{equation}
The step size $\beta_m$ may be fixed or selected through a line search that minimizes the loss along the direction given by $h_m$~\citep{boostingAlg_hothorn_2007}. This procedure is repeated for $m=1,\ldots,M$, where $M$ is the number of boosting stages. The number of stages and other hyperparameters can be selected using validation data or early stopping.

\begin{figure*}[t!]
\includegraphics[width=165mm]{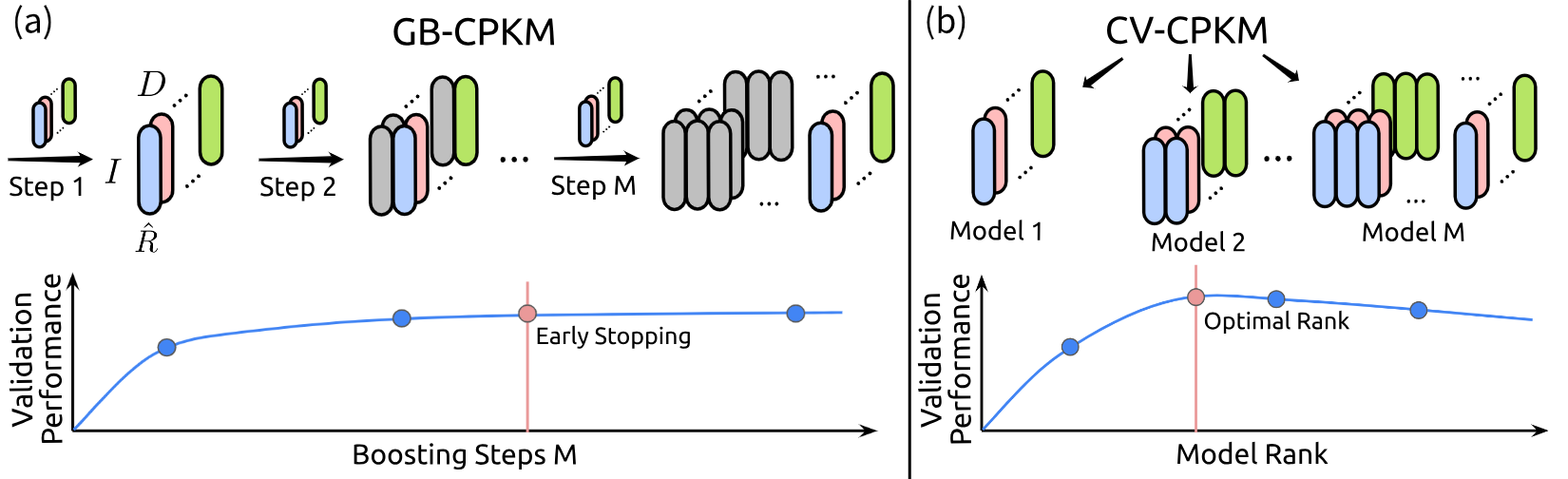}
\caption{Comparison of the training procedures for CP tensor network kernel machines (CPKMs) using gradient boosting (GB) and rank selection via validation (CV). (a) GB-CPKM incrementally builds an ensemble by training fixed-rank $\hat{R}$ base learners and concatenating their factor matrices along the rank dimension to represent the combined model. The ensemble size $M$ is selected based on validation performance. (b) In contrast, CV-CPKM independently trains candidate models with different CP ranks and selects the rank that achieves the lowest validation error. In the diagrams, each base learner model is represented by $D$ factor matrices of size $I \times \hat{R}$, where $I$ is the number of rows and $\hat{R}$ is the CP rank. Different colors distinguish the trainable factor matrices. Factors in gray are fixed and are not updated.}
\label{gb_diagram}
\end{figure*}

\subsection{CP Tensor Network Kernel Machine as a Base Learner}
In general, base learners can take many forms, including decision trees~\citep{addRegTree_friedman_2003} and neural networks~\citep{gbnn_badirli_2020}. In our approach, each base learner is represented as a CP kernel machine (CPKM) from Theorem~\ref{CPD_km}:
\begin{equation}
\label{tnkm_base}
    h_m(\vect{x})
    =
    \vect{\phi}(\vect{x})^\top
    \vect{w}(\vect{v}_m),
\end{equation}
where $\vect{w}(\vect{v}_m)$ is represented by a low-rank CPD parameterized by $\vect{v}_m$. For a rank-$\hat{R}$ CP decomposition, the parameter vector $\vect{v}_m$ has $\hat{R}\sum_{d=1}^{D} I_d$ entries. 

Substituting~\eqref{tnkm_base} into the update rule~\eqref{gb_update} and expanding the recursion yields the final ensemble model:
\begin{equation}
\label{gb_weight_sum}
    f_M(\vect{x})
    = \vect{\phi}(\vect{x})^\top \vect{w}(\hat{\vect{v}}) = 
    \vect{\phi}(\vect{x})^\top
    \left(
        \sum_{m=1}^{M}
        \beta_m \vect{w}(\vect{v}_m)
    \right).
\end{equation}
A key property of CP representations is that sums of CP tensors can be represented by concatenating their factor matrices $\mat{V}^{(d)}$, with the corresponding weights absorbed into the factors. Consequently, when each base learner has rank $\hat{R}$, the ensemble in~\eqref{gb_weight_sum} admits a representation with rank at most $M\hat{R}$:

\begin{equation*}
\label{higher_rank}
    \vect{w}(\hat{\vect{v}})
    = \sum_{m=1}^{M} \beta_m \vect{w}(\vect{v}_m) 
    = \sum_{(rm)=1}^{M\hat{R}} (\beta_m\vect{v}_m)^{(D)}_{r}
    \otimes \cdots \otimes
    (\vect{v}_m)^{(1)}_{r}.
\end{equation*}

For rank-one base learners, each boosting stage adds at most one CP component. Therefore, a rank-$R$ CPKM can be interpreted as a committee of $R$ rank-one learners. This perspective provides a way to adapt model complexity during training: each boosting stage adds new components, and the number of stages determines an upper bound on the ensemble's CP rank. By monitoring validation performance and stopping when further boosting no longer improves generalization, the ensemble size - and hence its effective rank - can be selected automatically alongside the base-learner parameters. This provides a practical alternative to explicitly searching over candidate ranks. The main differences between the two training strategies are illustrated in Figure~\ref{gb_diagram}.

The gradient boosting procedure for the squared-error loss is
summarized in Algorithm~\ref{alg:gb}. At each stage, a CPKM is
fitted to the residuals of the current ensemble, and its contribution is determined by a least-squares optimal step size $\beta_m$. The ensemble predictions are then updated by adding the new base learner.

{\setlength{\algomargin}{1.5em}
\begin{algorithm}[!t]
\DontPrintSemicolon
\KwParams{Number of boosting stages $M$, base-learner CP rank $\hat{R}$, regularization parameter $\gamma$, feature map $\vect{\phi}(\cdot)$}
\KwData{Training inputs $\mat{X} \inRe{N\times D}$ and targets
$\vect{y} \inRe{N}$}

\BlankLine

$\vect{f}_0 \gets \vect{0}$\;

\For{$m = 1,\ldots,M$}{
    \tcc{1. Compute residuals of the current ensemble}
    $\vect{r}_m \gets \vect{y}-\vect{f}_{m-1}$\;

    \tcc{2. Fit a regularized CPKM base learner to the residuals, using \eqref{cpd_km_opt}}
    $\displaystyle
    \vect{v}_m \gets
    \arg\min_{\vect{v}}
    \left[
    \frac{1}{2}
    \left\|\vect{r}_m-\mat{\Phi}\vect{w}(\vect{v})\right\|_2^2
    +
    \frac{\gamma}{2}
    \left\|\vect{v}\right\|_2^2
    \right]$\;

    \tcc{Compute the fitted base learner's predictions}
    $\vect{h}_m \gets \mat{\Phi}\vect{w}(\vect{v}_m)$\;

    \tcc{3. Compute the least-squares optimal step size}
    $\displaystyle
    \beta_m \gets
    \frac{\vect{r}_m^\top\vect{h}_m}
    {\vect{h}_m^\top\vect{h}_m}$\;

    \tcc{4. Update the ensemble predictions}
    $\vect{f}_m \gets \vect{f}_{m-1}+\beta_m\vect{h}_m$\;
}

\KwOutput{Base learner parameters
$\{\vect{v}_m\}_{m=1}^{M}$ and step sizes
$\{\beta_m\}_{m=1}^{M}$}
\caption{Gradient Boosting CPKM with Squared-Error Loss}
\label{alg:gb}
\end{algorithm}
}

\paragraph{Computational considerations.}
Let $E$ denote the number of ALS sweeps per base learner, $N$ is the number of training samples, $D$ is the input dimensionality, and $I$ is the number of basis functions per dimension. We assume that each base learner has a fixed CP rank $\hat{R}$.

\textit{Memory complexity.}
When $N \gg D$, the peak memory required to train a CPKM of CP
rank $R$ using ALS is $\mathcal{O}(NIR)$. In gradient boosting,
each base learner has a fixed CP rank $\hat{R}$ and is trained
sequentially. Therefore, the peak memory required for base-learner
training remains $\mathcal{O}(NI\hat{R})$, independently of the
number of boosting stages $M$, assuming that intermediate training
memory can be released between stages. In contrast, training a
single CPKM with CP rank $M\hat{R}$ requires
$\mathcal{O}(NIM\hat{R})$ peak memory under the same complexity
model. Although the memory required to train each base learner
remains bounded by its fixed rank, the memory required to store
the accumulated ensemble grows with $M$.

\textit{Training complexity.}
Under the assumed ALS implementation, the computational cost of
training a rank-$R$ CPKM is
$\mathcal{O}(ENDI^2R^2)$. Thus, training $M$ rank-one base learners
requires a cumulative cost of
$\mathcal{O}(ENDI^2M)$, assuming a comparable number of ALS sweeps
per learner. By comparison, training a single rank-$M$ CPKM has
complexity $\mathcal{O}(ENDI^2M^2)$ under the same complexity model.
This suggests a potential computational advantage for gradient
boosting, although the realized speedup depends on implementation
details and optimization behavior.

Overall, GB-CPKM provides a flexible way to increase model capacity incrementally using fixed-rank learners, potentially reducing training costs while enabling validation-based selection of the ensemble size. However, its sequential training procedure limits parallelism across boosting stages, and validation data are needed to select the stopping point and tune hyperparameters, such as the regularization strength.
\section{Experiments}
All experiments were conducted on a Dell Latitude 7440 laptop equipped with a 13th-generation Intel Core i7-1365U processor and 16~GB of RAM. We implement the proposed GB-CPKM Algorithm~\ref{alg:gb} from scratch, using the existing TNKM library~\citep{tnkm_2026_saiapin} for the underlying tensor network operations and model optimization. The Python source code, datasets, and experimental configurations required to reproduce the experiments are available on GitHub.\footnote{\url{https://github.com/AlbMLpy/boosting-tnkm}}

\subsection{Experimental Setup}

\paragraph{Datasets and Preprocessing.}
We evaluate the proposed GB-CPKM method on seven real-world regression datasets with varying characteristics and scales. Four datasets---\texttt{Energy}, \texttt{Concrete}, \texttt{Naval}, and \texttt{HouseElectric}---are obtained from the UCI Machine Learning Repository~\citep{uci_dua_2017}. We additionally consider \texttt{Kin8nm}~\citep{Kin8nm_Ghahramani_1996}, \texttt{Precipitation}~\citep{log_determinant_dong_2017}, and \texttt{Airline}~\citep{GPbigdata_Hensman_2023}.

For each dataset, the input features are scaled to the unit hypercube, and the target values are standardized. Each dataset is divided into training, validation, and test sets using a 70\%/10\%/20\% split, respectively. We use ten fixed random seeds to generate ten data splits. For each split, all the compared methods use identical training, validation, and test sets, enabling a paired comparison of predictive performance.

\paragraph{Model Configuration.} The primary objective is to evaluate the proposed GB-CPKM method against a cross-validation-based baseline (CV-CPKM), with particular emphasis on a CP rank selection and computational efficiency. Both methods use polynomial features~\citep{QTNM_Wesel_2024}. For simplicity, the number of basis functions is held constant across input dimensions, such that $I_d = I$ for all $d=1, \dots, D$. The basis size $I$ is selected separately for each dataset to provide sufficient model capacity while keeping memory requirements within the available hardware constraints. For a given dataset, the same basis size is used for both methods to ensure a consistent representation. We use the alternating least squares (ALS) algorithm to train the tensor network models and individual gradient-boosting base learners for five epochs. Regularization is applied to the CP cores. The regularization coefficient $\gamma$ is selected from the candidate set \[ \gamma \in \{10^{-6}, 10^{-5}, 10^{-4}, 10^{-3}, 10^{-2}\}. \] Both GB and CV tune the regularization coefficient using the validation set.

\paragraph{Evaluation Metrics and Computational Cost.} We evaluate both methods on the test set for each of the ten data splits and report the mean and standard deviation of the resulting mean squared error (MSE). We additionally measure training time in seconds and peak memory consumption in mebibytes (MiB).

\subsection{Effect of Base-Learner Rank}
The first set of experiments investigates how the rank of individual base learners and the number of boosting stages affect the predictive performance and computational efficiency of GB-CPKM. Our aim is to understand how a fixed model capacity should be distributed between the complexity of individual learners and the number of learners in the ensemble. 
To this end, we consider base-learner ranks $R \in \{1,2,4,8,12\}$ and fix the total nominal rank budget at $R_{\mathrm{total}}=24$. Accordingly, the number of boosting stages is set to $N_{est}=24/R$, yielding $N_{est} \in \{24,12,6,3,2\}$, respectively. This setup allows us to compare ensembles comprising many low-rank learners with ensembles comprising fewer higher-rank learners, while keeping the total nominal rank budget constant.

\begin{figure*}[t!]
\includegraphics[width=165mm]{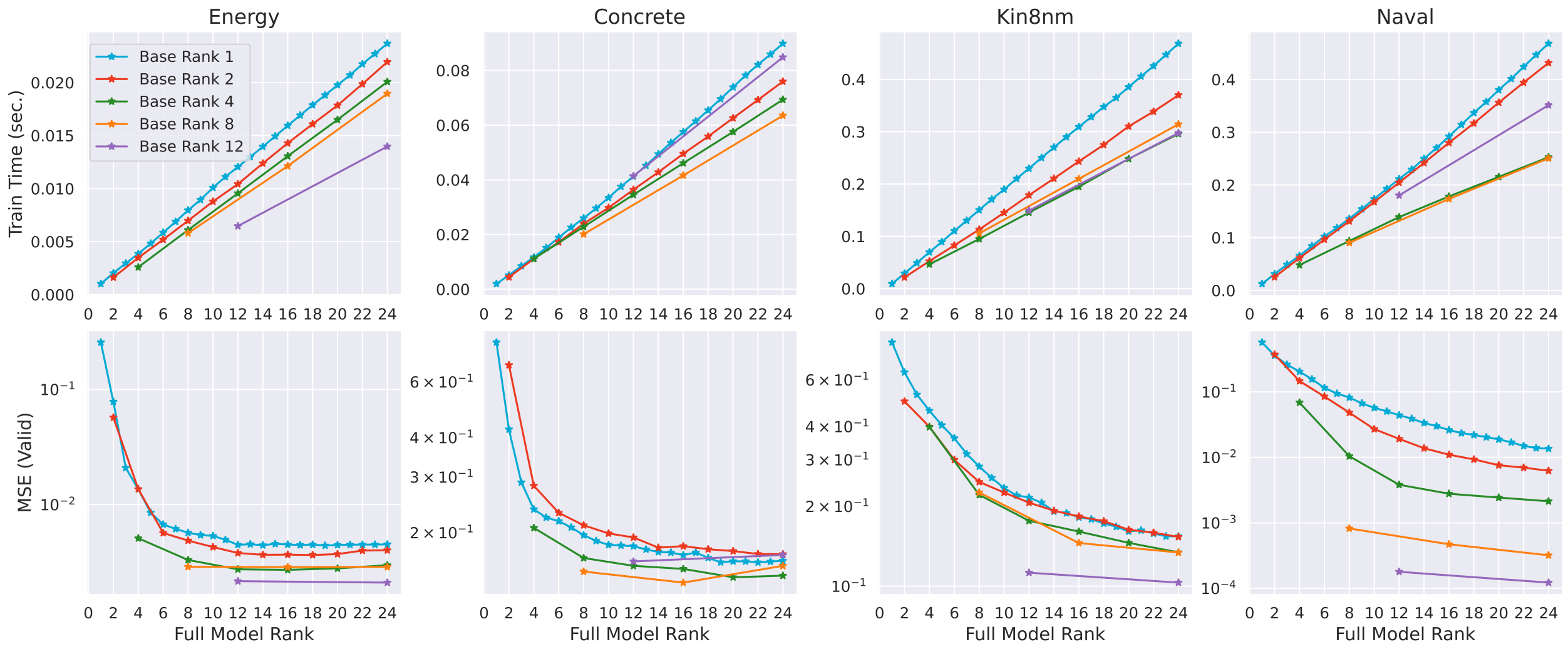}
\caption{Ablation study of base-learner rank in GB-CPKM on four real-world datasets, with each column corresponding to a dataset. The top row shows cumulative training time (in seconds), while the bottom row reports validation MSE on a logarithmic scale. Across the evaluated datasets, higher-rank base learners achieve better predictive performance while also reducing cumulative training time. However, rank selection should also account for memory requirements, which may constrain the use of higher-rank models in practice.}
\label{base_rank_ablations}
\end{figure*}

Figure~\ref{base_rank_ablations} presents the cumulative training time and validation MSE for the four datasets. The results indicate that increasing the base-learner rank generally improves predictive performance. Higher-rank configurations also tend to reduce cumulative training time, particularly when comparing base ranks 4, 8, and 12, although their relative training times vary across datasets. These findings suggest that allocating a fixed rank budget to fewer, higher-rank learners can be more effective than distributing it across a larger number of lower-rank learners, improving predictive accuracy while potentially reducing computational cost.

\subsection{Gradient Boosting for Efficient Rank Search}
In this section, we compare GB-CPKM with a cross-validation (CV) baseline for selecting CP ranks in a sequential learning setting. The goal is to investigate whether gradient boosting with fixed-rank base learners can provide an effective alternative to explicitly searching over candidate ranks.

GB-CPKM uses base learners with a fixed CP rank of $2$. We use ten boosting stages for all datasets except \texttt{Airline}, for which we use five. In contrast, the CV-CPKM baseline evaluates a predefined set of candidate ranks. For all datasets except \texttt{Airline}, the candidate set is $ R \in \{2,4,6,\ldots,20\}$, whereas for \texttt{Airline}, we use $ R \in \{2,4,6,8,10\}$. For each candidate rank, the model is trained on the training set and evaluated on the validation set. The rank achieving the lowest validation MSE is selected. The test set is reserved exclusively for final evaluation and is not used for hyperparameter selection.

\subsubsection{Dynamical Behavior on Small-Scale Problems}
In this experiment, we investigate how increasing the number of CP rank options affect the test dynamics of CPKM under the two learning strategies, GB and CV, on small-scale problems. The goal is to characterize the trade-offs between sequentially adding fixed-rank base learners and explicitly evaluating candidate ranks during hyperparameter selection.

\begin{figure*}[t!]
\includegraphics[width=165mm]{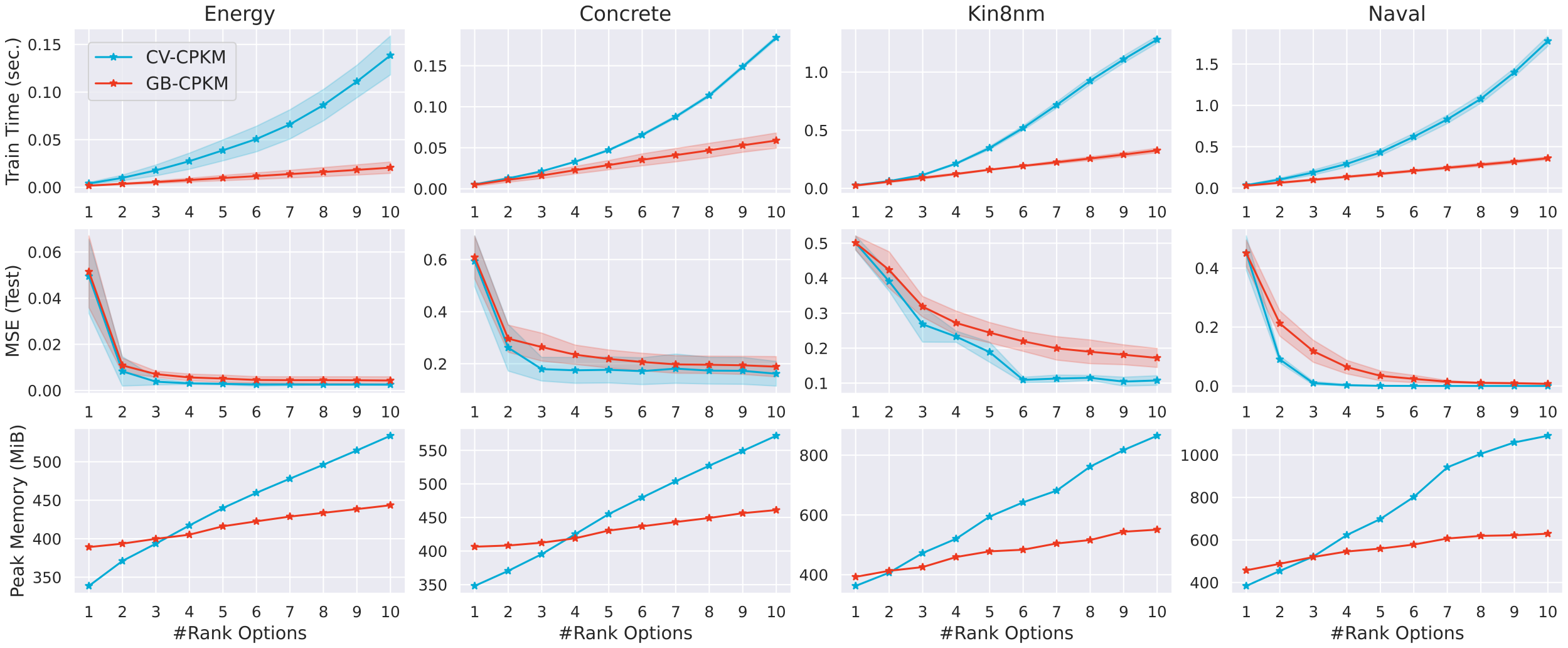}
\caption{Comparison of GB-CPKM (red) and CV-CPKM (blue) on four real-world datasets, with each column corresponding to a dataset. From top to bottom, the rows show cumulative training time (in seconds), test MSE, and peak memory consumption (in MiB), plotted against the number of considered candidate CP ranks. Solid lines indicate the mean across 10 runs, and shaded regions represent $\pm$ 1 standard deviation. GB-CPKM consistently requires less training time and peak memory than CV-CPKM, while achieving comparable predictive performance. The test MSE of GB-CPKM is similar to that of CV-CPKM, with overlapping variability bands in some cases, and approaches comparable values across the evaluated datasets. These results highlight the potential of GB-CPKM to reduce computational costs without substantially compromising predictive accuracy.}
\label{rank_dynamics}
\end{figure*}

Figure~\ref{rank_dynamics} shows how test MSE, cumulative training time, and peak memory consumption change as the number of candidate rank options increases. Across all datasets, GB-CPKM exhibits a slower increase in cumulative training time than cross-validation baseline. This advantage stems from its sequential training strategy, which reuses previously trained base learners, whereas CV-CPKM trains a separate model from scratch for each candidate rank. The same difference in training strategies also helps explain the lower peak memory consumption of GB-CPKM. Since it uses fixed-rank base learners, the memory required to train each learner and the whole model is primarily determined by the base rank. In contrast, CV-CPKM evaluates candidate models of increasing rank, requiring progressively more memory as model complexity grows. Consequently, its peak memory consumption is influenced by the largest candidate rank considered, which can lead to substantially higher memory requirements.

Speaking of predictive performance, GB-CPKM achieves test MSE values comparable to those of CV-CPKM on most datasets. The mean test errors are similar, and their standard-deviation bands overlap in several cases, suggesting that GB-CPKM can achieve competitive predictive accuracy at a lower computational cost. However, these observations alone do not establish statistical equivalence between the methods.

\subsubsection{Large-Scale Problems}
In this experiment, we further evaluate the gradient boosting model on three large-scale datasets to investigate its scalability, while also including results for smaller datasets to provide a broader view of its performance across different problem sizes. 

In Table~\ref{table:hypers-search} we present the comparison of GB-CPKM with the cross-validation baseline (CV-CPKM) with a focus on the trade-offs between predictive performance, training time (rank search), and peak memory consumption. The results show that GB model offers clear computational advantages over cross-validation, training up to $6\times$ faster and requiring up to $2\times$ less peak memory. In terms of predictive performance, the two methods achieve broadly comparable test MSE, although GB-CPKM yields somewhat higher errors on some datasets, for example Naval. 

We also introduce a hybrid variant, GB-CPKM(H), which follows the same training procedure as GB-CPKM but includes one additional full ALS sweep over the combined model before test evaluation. This additional optimization considerably improves predictive performance on most datasets. However, optimizing the full model increases peak memory consumption to levels comparable to those of CV-CPKM, reducing the memory-efficiency advantage of the original model.

In summary, these results suggest that gradient boosting provides a more computationally efficient alternative to explicit rank search, substantially reducing training time and memory requirements while maintaining competitive predictive performance on the evaluated large-scale problems.

\begin{table}
\caption{Performance comparison of GB-CPKM and CV-CPKM across regression benchmarks. We report test mean squared error (MSE), training time (seconds), and peak memory consumption (MiB). Each experiment is repeated over 10 runs, with results reported as mean $\pm$ 1 standard deviation. $N$ denotes the number of samples, and $D$ denotes the input dimensionality. Overall, GB-CPKM trains up to $6\times$ faster and uses up to $2\times$ less peak memory, on average, than CV-CPKM, while achieving comparable test MSE. The
hybrid variant, GB-CPKM(H), further improves predictive performance
on most datasets by performing an additional full ALS sweep on the
combined model.}
\label{table:hypers-search}
\begin{center}
\scalebox{0.74}{
\begin{tabular}{||l||c|c||c|c|c||c|c||c|c||}
\toprule
 &  &  & \multicolumn{3}{c||}{MSE (test) $\downarrow$} & \multicolumn{2}{c||}{Time (sec.) $\downarrow$} & \multicolumn{2}{c||}{Memory (MiB) $\downarrow$} \\
\midrule
 Data & N & D & CV-CPKM & GB-CPKM & GB-CPKM(H) & CV-CPKM & GB-CPKM & CV-CPKM & GB-CPKM \\
\midrule
Energy & 768 & 8 & \textbf{0.0026}±0.00 & 0.0043±0.00 & 0.0038±0.00 & 0.14±0.02 & \textbf{0.02}±0.01 & 534 & \textbf{444} \\
Concrete & 1030 & 8 & \textbf{0.1618}±0.05 & 0.1889±0.04 & 0.1801±0.05 & 0.18±0.00 & \textbf{0.06}±0.01 & 572 & \textbf{461} \\
Kin8nm & 8192 & 8 & \textbf{0.1077}±0.01 & 0.1721±0.03 & 0.1209±0.01 & 1.28±0.03 & \textbf{0.33}±0.02 & 865 & \textbf{551} \\
Naval & 11934 & 16 & \textbf{0.0000}±0.00 & 0.0074±0.00 & 0.0010±0.00 & 1.78±0.06 & \textbf{0.36}±0.02 & 1090 & \textbf{630} \\
\midrule
Precipitation & 628474 & 3 & \textbf{0.9419}±0.04 & 0.9454±0.04 & 0.9448±0.04 & 88.59±11.84 & \textbf{8.35}±1.05 & 2885 & \textbf{1156} \\
HouseElectric & 2049280 & 6 & \textbf{0.0010}±0.00 & 0.0018±0.00 & \textbf{0.0010}±0.00 & 269.60±12.67 & \textbf{25.21}±0.58 & 6058 & \textbf{2214} \\
Airline & 5929413 & 8 & \textbf{0.8272}±0.06 & 0.8607±0.04 & 0.8395±0.03 & 196.86±7.65 & \textbf{45.01}±3.07 & 9802 & \textbf{4023} \\
\bottomrule
\end{tabular}
}
\end{center}
\end{table}
\section{Related Work}
Selecting an appropriate tensor rank remains a challenging problem in tensor decomposition and tensor-network learning. A common practice is to treat rank as a hyperparameter and select it using grid search or cross-validation over a predefined set of candidate values~\citep{tdspml_sidiropoulos_2017, em_novikov_2017, fftd_wesel_2021, slnsd_haliassos_2021, tnsrkf_menzen_2026}. Although conceptually simple, these approaches require training and evaluating multiple models, potentially increasing computational and memory costs.

\paragraph{Bayesian approaches.} Several studies address rank determination through probabilistic modeling and sparsity-inducing priors. \citet{bstucker_zhao_2015} introduce probabilistic Tucker models that infer multilinear rank using structured sparsity and model evidence. \citet{btbr_camarrone_2016} propose Bayesian $N$-way partial least squares with automatic relevance determination (ARD) priors on CP factors to infer the number of relevant components. More recently, \citet{btnkm_kilic_2025} develop a probabilistic framework for tensor network kernel machines that uses hierarchical sparsity-inducing priors to infer CP rank and feature dimensions. These methods offer principled mechanisms for automatic complexity determination, but require probabilistic formulations and associated inference procedures.

\paragraph{Adaptive tensor-network optimization.} Rank adaptation has also been studied in tensor-network optimization, particularly in the tensor-train (TT), or matrix product state (MPS), format. \citet{alstt_holtz_2012} develop ALS- and MALS-based optimization methods that allow TT ranks to adapt during optimization. Related strategies appear in density matrix renormalization group (DMRG) algorithms~\citep{dmrg_schollwöck_2011}. Similarly, \citet{amen_dolgov_2014} propose methods for solving high-dimensional symmetric positive-definite linear systems using low-rank tensor representations and sequential optimization of tensor components. While these methods demonstrate the potential of adaptive rank optimization, they primarily focus on the tensor-train (TT) format, whereas our work investigates rank allocation in canonical polyadic (CP) tensor-network models.

\paragraph{Reinforcement learning.} \citet{cprank_zhang_2024} formulate CP rank selection as a decision-making problem using the Deep Deterministic Policy Gradient (DDPG) algorithm. Their method combines a pretrained model of tensor information with an SVD-based exploration strategy to select ranks for convolutional-layer compression. This provides a learned alternative to conventional rank search, but is designed for a specific application and requires a reinforcement-learning framework. 

\paragraph{Positioning of our work.} Rather than introducing a separate probabilistic inference procedure, a reinforcement-learning agent, or an adaptive-rank optimization algorithm, we investigate gradient boosting with fixed-rank CP tensor-network base learners. GB-CPKM incrementally increases model capacity by adding learners, providing a straightforward alternative to explicit rank search. A comprehensive comparison with Bayesian and reinforcement-learning methods is beyond the scope of this study, as these approaches involve different modeling assumptions and computational procedures.
\section{Conclusion}
We introduced a new gradient boosting model, GB-CPKM, that incrementally increases model capacity using fixed-rank CPKM base learners, providing an alternative to resource-intensive
CP rank hyperparameter search. Experiments on real-world regression datasets of varying sizes and dimensionalities show that GB-CPKM trains up to $6\times$ faster and uses up to $2\times$ less peak memory than a cross-validation-based rank-selection baseline, while achieving comparable predictive performance. The hybrid variant GB-CPKM(H) further improves predictive performance on most datasets through an additional ALS sweep over the combined model, though at the cost of increased peak memory consumption.

Several directions remain for future work. We plan to extend the framework to tensor train kernel machines (TTKMs) and investigate rank adaptation strategies that allow ranks to increase or decrease during training. Another promising direction is to explore broader applications of GB-CPKM beyond regression, including classification and time-series forecasting.

\subsubsection*{Acknowledgments}
This publication is part of the project Sustainable learning for Artificial Intelligence from noisy large-scale data (with project number VI.Vidi.213.017) which is financed by the Dutch Research Council (NWO).

\bibliography{main}
\bibliographystyle{tmlr}

\end{document}